\documentclass{article}

\usepackage{microtype}
\usepackage{graphicx}
\usepackage{subfigure}
\usepackage{booktabs} 
\usepackage{multirow} 
\usepackage{hyperref}

\usepackage[accepted]{icml2024}

\usepackage{amsmath}
\usepackage{amssymb}
\usepackage{mathtools}
\usepackage{amsthm}

\usepackage[capitalize,noabbrev]{cleveref}

\theoremstyle{plain}

\theoremstyle{definition}

\theoremstyle{remark}

\newcommand{\lms}[1]{{\textcolor{black}{#1}}}
\newcommand{\cky}[1]{{\textcolor{black}{#1}}}
\newcommand{\gxz}[1]{{\textcolor{black}{#1}}}

\usepackage[textsize=tiny]{todonotes}

\icmltitlerunning{CogDPM: Diffusion Probabilistic Models via Cognitive Predictive Coding}

\begin{document} 

\twocolumn[

\icmltitle{CogDPM: Diffusion Probabilistic Models via Cognitive Predictive Coding}

%


\icmlsetsymbol{equal}{*}

\begin{icmlauthorlist}
\icmlauthor{Kaiyuan Chen}{equal,sch}
\icmlauthor{Xingzhuo Guo}{equal,sch}
\icmlauthor{Yu Zhang}{sch}
\icmlauthor{Jianmin Wang}{sch}
\icmlauthor{Mingsheng Long}{sch}
\end{icmlauthorlist}

\icmlaffiliation{sch}{School of Software, BNRist, Tsinghua University. Kaiyuan Chen $<$cky21@mails.tsinghua.edu.cn$>$}

\icmlcorrespondingauthor{Mingsheng Long}{mingsheng@tsinghua.edu.cn}

\icmlkeywords{Diffusion model, Predictive coding, Deep learning}

\vskip 0.3in
]



\printAffiliationsAndNotice{\icmlEqualContribution} 



\begin{abstract}
Predictive Coding (PC) is a theoretical framework in cognitive science suggesting that the human brain processes cognition through spatiotemporal prediction of \gxz{the} visual world. Existing studies have developed spatiotemporal prediction neural networks based on the PC \lms{theory}, emulating its two core mechanisms: Correcting predictions from residuals and hierarchical learning. However, these models do not show the enhancement of prediction skills on real-world forecasting tasks and ignore the \textit{Precision Weighting} mechanism of PC theory. \gxz{The} precision \lms{weighting} mechanism posits that the brain allocates more attention to signals with lower precision, contributing to the cognitive ability of human brains. This work introduces the \textit{Cognitive Diffusion Probabilistic Models} (CogDPM)\gxz{,} which demonstrate the connection between diffusion probabilistic models and PC theory. CogDPM features a precision estimation method based on the hierarchical sampling capabilities of diffusion models and weight the guidance with precision weights estimated by the inherent property of diffusion models. We experimentally show that the precision \lms{weights} effectively estimate the data predictability. We apply CogDPM to real-world prediction tasks using the United Kindom precipitation and ERA surface wind datasets. Our results demonstrate that CogDPM outperforms both existing domain-specific operational models and general deep prediction models \lms{by} providing more proficient forecasting. 
\end{abstract}

\section{Introduction}
\label{Introduction}

Predictive Coding (PC) is a theoretical construct in cognitive science, positing that the human brain cognizes the visual world through predictive mechanisms~\cite{spratling2017review,hohwy2020new}. The PC theory elucidates that the brain hierarchically amends its perception of the environment by anticipating changes in the visual world. Researchers have developed computational models based on the PC theory to simulate the brain's predictive mechanisms~\cite{keller2018predictive}. Neuroscientists employ these models to empirically validate the efficacy of the PC theory and to find new characteristics. Precision weighting, a pivotal feature of the PC theory, suggests that the brain assigns more attention to signals with lower precision by using precision as a filter in weighting prediction errors.

With the advancement of deep learning, predictive learning has emerged as one of the principal learning methods ~\cite{rane2020prednet,bi2022pangu}. Neural networks are now capable of making effective predictions in video data~\cite{shi2015convolutional,wang2017predrnn,ho2022video}. Deep video prediction models have rich applications, such as weather forecasting~\cite{ravuri2021skilful,zhang2023skilful} and autonomous driving simulation~\cite{wang2018video,wen2023panacea}.

Researchers design cognitively inspired video prediction models utilizing the PC theory. PredNet~\cite{lotter2020neural}, which employs multi-layer ConvLSTM~\cite{shi2015convolutional} networks to predict the next frame in a video sequence, is responsible for predicting the residual between the outcomes of a network layer and the ground truth values. However, \gxz{the} predictive capability of PredNet does not show significant improvement over non-hierarchical video prediction models and has not been validated in real-world video prediction tasks. We posit that the hierarchical modeling mechanism in PredNet is not effectively implemented. PredNet directly targets low signal-to-noise ratio residuals as learning objectives, which complicates the learning process, and fails to extract fundamentally distinct features between layers. Additionally, PredNet lacks the capability to model precision, leading to uniform weighting in learning residuals across different regions. This results in redundant noise information becoming a supervisory signal and hinders the model's ability to learn from important information.

In this study, we propose PC-inspired \emph{Cognitive Diffusion Probabilistic Models} (CogDPM), which align the main features of PC theory with Diffusion Probabilistic Models (DPMs), a specialized branch of deep generative models.
The CogDPM framework innovatively abstracts the multi-step inference process characteristic of Diffusion Probabilistic Models into a hierarchically structured model, where each layer is responsible for processing signals at distinct spatiotemporal scales. 
This hierarchical approach allows for a progressive enhancement in the model's interpretation of sensory inputs, actively working to reduce prediction errors through iterative refinement. 
A key feature of the CogDPM framework is its ability to estimate spatiotemporal precision weights based on the variance of states in each hierarchical layer. 
This methodology plays a crucial role in optimizing the overall precision of predictions, and represents a novel advancement in predictability modeling.

We verify the effectiveness of precision weights as well as the predictions skills of CogDPM on real-world spatiotemporal forecasting tasks.
To verify precision weights, we use synthetic motion datasets of both rigid body and fluid. Results show precision weights get higher salience on the hard-to-predict region.
To validate the prediction capabilities of CogDPM, we apply CogDPM to real-world tasks including precipitation nowcasting~\cite{shi2015convolutional,ravuri2021skilful} and high wind forecasting~\cite{barbounis2006long,soman2010review}. 
We evaluate CogDPM through case studies focusing on extreme weather events and scientific numerical metrics. CogDPM outperforms operational domain-specific models FourCastNet~\cite{pathak2022fourcastnet} and DGMR~\cite{ravuri2021skilful} as well as the general deep predictive models. 
We demonstrate that CogDPM has strong extreme event prediction capabilities and verify the effectiveness of precision estimations of CogDPM which provide useful information for weather-driven decision-making.

In summary, we identify the following advantages of CogDPM:
\begin{itemize}
    \item CogDPM aligns diffusion probabilistic models with Predictive Coding theory, which inherently integrates hierarchy prediction error minimization with precision-weighting mechanics.

    \item CogDPM delivers skillful and distinct prediction results, particularly in scientific spatiotemporal forecasting, demonstrating a marked improvement \gxz{in} probabilistic forecasting metrics.

    \item CogDPM presents a novel method for predictability estimation, providing index of confidence modeling for probabilistic forecasting.

\end{itemize}

\section{Related Work}

\paragraph{Predictive Learning.}
Predictive learning is a subfield of machine learning that utilizes historical data to make predictions about future events or outcomes.
\gxz{As} an important aspect of human cognition that plays a crucial role in our ability to perceive and understand the world, spatiotemporal predictive learning has triggered a substantial amount of research efforts, such as ConvLSTM~\cite{shi2015convolutional}, PredRNN~\cite{wang2017predrnn}, and ModeRNN~\cite{yao2021modernn}.
Recently, diffusion models~\cite{ho2020denoising} have been successfully applied in video generation~\cite{ho2022imagen} so as to capture spatiotemporal correlations, showing a promising trend as a spatiotemporal predictive learning framework.

\paragraph{Predictive Coding.} 
In neuroscience, predictive coding is a theory of brain function about how brains create predictions about the sensory input.
\citeauthor{rao1999predictive} translates the idea of predictive coding into a computational model based on extra-classical receptive-field effects, and shows the brain mechanism of trying to efficiently encode sensory data using prediction.
Further research in neuroscience~\cite{friston2009free,clark2013whatever,emberson2015top,spratling2017review} presents different interpretations of predictive coding theory.


\paragraph{Predictive Coding Neural Networks.}
The development of deep learning has arisen plenty of deep predictive networks with cognition-inspired mechanisms.
PredNet~\cite{lotter2016deep} implements hierarchical predictive error with ConvLSTM for spatiotemporal prediction using principles of predictive coding.
CPC~\cite{oord2018representation,henaff2020data} and MemDPC~\cite{han2020memory} incorporate contrastive learning in the latent space via a predictive-coding-based probabilistic loss.
PCN~\cite{wen2018deep,han2018deep} proposes a bi-directional and recurrent network to learn hierarchical image features for recognition.
Such models introduce the motivation of predictive coding in their task-specific manners. However, these works ignore precision weighting, a pivotal mechanism in PC theory. Besides, these works have not explored a proper PC-based framework of diffusion models.

\section{Method}\label{sec:method}

Spatiotemporal forecasting involves extracting patterns from a sequence of vector fields $\mathbf{c}^{-N_0:0}$ and providing future evolution $\mathbf{x}^{1:N}$. We give a brief introduction to the framework of predictive coding and propose our CogDPM for implementing Predictive Coding into spatiotemporal forecasting.

\cky{To avoid confusion, we use the superscript $N$ to represent different moments allowed by time, and the subscript $t$ to denote the ordinal number of the inference steps in the diffusion model.
}

\subsection{CogDPM via Predictive Coding}

\begin{figure*}[ht]
    \includegraphics[width=\textwidth]{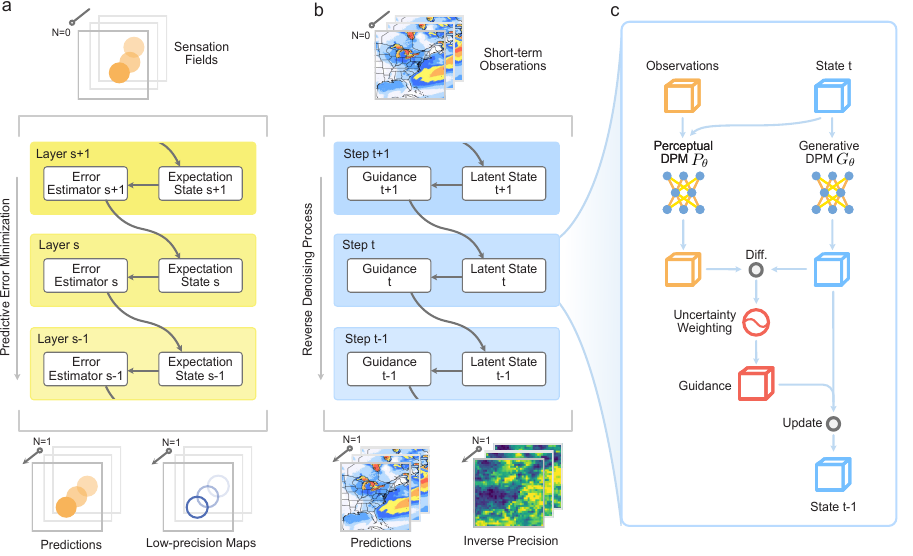}
    \vspace{-6mm}
    \caption{\textbf{a}, A general predictive coding framework. The system recognizes the sensation fields with hierarchy error units and expectation units and generates the predictions and precision maps during the process. \textbf{b}, Cognitive Diffusion Probabilistic Models (CogDPM) framework, providing predictions and precision weights with multi-step denoising process. \textbf{c}, Updates of latent states with precision-weighted predictive error.}
    \label{fig:NPP}
    \vspace{-6pt}
\end{figure*}

Figure~\ref{fig:NPP}a presents a conceptual demonstration of a predictive coding (PC) system. Based on PC theory, we propose \emph{Cognitive Diffusion Probabilistic Models (CogDPM)} for spatiotemporal forecasting based on multi-step denoising~\cite{ho2020denoising}, which realizes the core mechanisms of hierarchical inference and prediction error minimization. Fig.~\ref{fig:NPP}b shows the framework of CogDPM, which takes past observations as input to forecast the evolution of future fields and estimate corresponding prediction error.

\paragraph{Hierarchical Inference.}
Predictive coding theory describes that the brain makes spatiotemporal predictions of the sensations through hierarchical inference with multi-layer organized estimators~\cite{walsh2020evaluating}. 
While different layers of the PC system are responsible for processing features at different spatial scales, the hierarchical system gradually performs prediction error minimization and converges on a final consistent predictions~\cite{wiese2017vanilla}.
CogDPM aligns the multi-step inference \lms{of DPM} with the hierarchical inference of the PC system. 
In the inference phase of CogDPM, the forecast is gradually generated in the hidden states evolution process from $\mathbf{x}_T,\mathbf{x}_{T-1},\dots$ to $\mathbf{x}_0$, where $\mathbf{x}_T$ is a Gaussian prior and $\mathbf{x}_0$ \lms{indicates} the generated target distribution of forecast. 
CogDPM inherits the properties of DPM that the different inference steps have varying spatial and temporal scales of feature expression capabilities~\cite{zheng2022entropy}. In the initial stages of inference, the model yields holistic and vague results. As it approaches the final steps, the model shifts its focus towards supplementing with detailed information, which is also aligned with the hierarchical property of the PC system. 
In each internal inference step, the guidance of the diffusion model plays a similar role with the error units \lms{of the PC system}, taking observation sequence as input and strengthen the correlation between generated results and observations~\cite{dhariwal2021diffusion}.

\paragraph{Prediction Error Minimization.} 
Each layer in the PC system outputs two key components: predictions for future sensations and estimations of prediction errors~\cite{van_elk_2021predictive}. 
This process is enabled by interactions between two functionally distinct neural sub-components in the layer: expectation units and error units~\cite{walsh2020evaluating}.
The expectation unit updates expected sensory states from the previous level to the error units, without directly receiving sensory-driven signals as input.
The error unit receives and analyzes the discrepancies between perceptual and expected sensory states to compute the error, which is then fed back to the expectation unit in the next layer.
The goal of the information transfer between multiple layers is to minimize prediction errors, ultimately resulting in more accurate environmental perceptions.
CogDPM couples a generative DPM $G_{\theta}$ with a perceptual DPM $P_{\theta}$, where $\theta$ \lms{represents} their sharing parameters. 
The previous state $\mathbf{x}_{t}$ is the sharing input of both models, while observations $\mathbf{c}$ can only be attached by the perceptual DPM.
With the previous state as observation, the perceptual DPM acts as sensory stimuli and thus aligns with the bottom-up process in \lms{the PC system}.
The generative DPM, as a \lms{comparison}, performs as the top-down prediction based on conceptual knowledge.
%
Fig.~\ref{fig:NPP}c provides detailed schematic diagram of a single step in CogDPM.
Given the outputs $G_{\theta}(\mathbf{x}_{t})$ and $P_{\theta}(\mathbf{x}_{t}, \mathbf{c})$ separately for each step $t$, the guidance for predictive error minimization can be expressed by:
\begin{align}
\text{Guidance}[\mathbf{x}_t] = P_{\theta}(\mathbf{x}_{t}, \mathbf{c}) - G_{\theta}(\mathbf{x}_{t})
\label{eq:error},
\end{align}
i.e., the difference between \lms{sensations} and predictions.

\subsection{Precision Weighting in CogDPM}
Precision weighting stands as the pivotal mechanism for filtering information transmitted between adjacent layers. It posits that the brain expends more effort in comprehending imprecise information, recognizing that sensory input often contains a substantial proportion of redundant information, which does not necessitate repetitive processing~\cite{hohwy2020new}. During each error minimization phase of the predictive coding (PC) approach, the error unit generates precision maps. These maps selectively filter the signal transmitted to the subsequent layer, assigning greater weight to signals characterized by higher imprecision. 

Following precision weighting in PC theory, our goal is to design a modeling of imprecision for each denoising process of CogDPM. We therefore delve into the progressive denoising mechanism in the backward process of DPMs. In each denoising step for $\mathbf{x}_t$, the model predicts a noise towards the corresponding groundtruth $\mathbf{x}_0$~\citep{song2020denoising}. 
%
The model usually shifts $\mathbf{x}_t$ into $\mathbf{x}_{t-1}$ within a tiny step and recursively \lms{performs} the process to get $\mathbf{x}_0$, but can either directly obtain $\mathbf{x}_0$ within a single larger step.
\gxz{If the direct predictions from step $t$ and from step ${t+1}$ with generative DPM $G_\theta$
differ in a significant manner for a certain spatiotemporal region,
the single step produces inconsistent signal from previous steps, indicating the imprecision of the generative model at such region of the current state.} Hence, we use the fluctuation field of direct predictions $\mathbf{x}_0$ from $\{\mathbf{x}_t, \dots, \mathbf{x}_{t+k-1}\}$ to estimate such imprecision of state $\mathbf{x}_t$ for each coordinate, formulated by Eq.~\eqref{eq:precision}:
%
%
%
\begin{align}
    \text{U}[\mathbf{x}_t] = \text{Var}\left[\mathbb{E}_{G_{\theta}}\left[\mathbf{x}_0 \mid \mathbf{x}_t\right],\dots, \mathbb{E}_{G_{\theta}}\left[\mathbf{x}_{0} \mid \mathbf{x}_{t+k-1} \right]\right]
\label{eq:precision},
\end{align}
\gxz{where $\mathrm{Var}$ stands for the variance field along the denoising step, and $k$ is the hyperparameter for window length.
In this way, CogDPM provides a modeling of the inverse precision field for multiscale spatiotemporal coordinates in the inference steps.
}
Since only the past observation is given in the forecasting tasks, this precision is a good substitution for the actual precision to weight the minimization.
We implement precision weighting in the CogDPM framework, which can be formulated as Eq.~\eqref{eq:Update-precision},
\begin{equation}
    \mathbf{x}_{t-1} = G_{\theta}(\mathbf{x}_{t}) + f(\text{U}[\mathbf{x}_t]) \cdot \text{Guidance}[\mathbf{x}_t]
    \label{eq:Update-precision},
\end{equation}
where $f$ is a parameter-free normalization function shown in Eq.~\eqref{eq:Post-operation}. 
%
Precision weighting helps to control the balance between diversity and the alignments with the observation, with larger guidance increasing the alignments and decreasing the diversity or the quality of generations. 
Through this precision weighting mechanism, CogDPM strategically allocates greater guidance intensity to regions with lower predictability, thereby enhancing local precision in a focused manner.

\paragraph*{Computational details.} \label{Extend:NPP}

The framework of a standard DPM starts with $\mathbf{x}_0$ sampled from data distribution, and latent states $\{\mathbf{x}_1,\ \mathbf{x}_2,\dots, \mathbf{x}_T\}$ following the forward process \lms{along a Markov chain} as Eq.~\eqref{eq:DPM-forward}.
\begin{align}
    q(\mathbf{x}_{t+1}\mid \mathbf{x}_{t}) = \mathcal{N}\left(\sqrt{\alpha_{t}}\mathbf{x}_{t}, \sqrt{1-\alpha_{t}} \mathbf{I}\right) 
    \label{eq:DPM-forward},
\end{align}
where $\{\alpha_t\}_{t=1, 2,\dots, T}$ are constant parameters.
Each latent state is a corrupted estimation for the future inputs with the three-dimensional shape of $N\times H\times W$.

In each step of the backward process, we update the latent state with the denoising network $\mathbf{\epsilon}_{\theta}$.  
We denote the sensation input as $\mathbf{c}$, which has a shape of $N_0\times H\times W$. The perceptual model $P_{\theta}$ and generative model $G_{\theta}$ can be preformed separately as Eq.~\eqref{eq:P-theta} and~\eqref{eq:G-theta}. 
\begin{align}
    \label{eq:P-theta}
    P_{\theta}(\mathbf{x}_t, \mathbf{c}) &= \frac{1}{\sqrt{\alpha}_t}\left(\mathbf{x}_t - \frac{1- \alpha_t}{\sqrt{1-\bar{\alpha}_t}}\mathbf{\epsilon}_{\theta}(\mathbf{x}_t, \mathbf{c})\right),\\
    \label{eq:G-theta}
    G_{\theta}(\mathbf{x}_t) &=\frac{1}{\sqrt{\alpha}_t}\left(\mathbf{x}_t - \frac{1- \alpha_t}{\sqrt{1-\bar{\alpha}_t}}\mathbf{\epsilon}_{\theta}(\mathbf{x}_t, \varnothing)\right),
\end{align}
where $\bar{\alpha}_t = \prod_{s=1}^t \alpha_s$ and $\epsilon_\theta$ is the denoising network of the DPM.
CogDPM \lms{provides} inverse precision estimation with Eq.~\eqref{eq:precision}, and $\mathbb{E}_{G_{\theta}}\left[\mathbf{x}_0\mid \mathbf{x}_t\right]$ can be \lms{computed} as Eq.~\eqref{eq:Reverse-s-to-S}:
\begin{equation}
    \mathbb{E}_{G_{\theta}}\left[\mathbf{x}_0\mid \mathbf{x}_t\right] = \frac{1}{\sqrt{\bar{\alpha}_t}}\left(\mathbf{x}_t - \sqrt{1-\bar{\alpha}_t}\epsilon_{\theta}(\mathbf{x}_t, \varnothing)\right).
    \label{eq:Reverse-s-to-S}
\end{equation}
For implementation, we push $\lms{G}_{\theta}\left(\mathbf{x}_t\right)$ into the estimation queue with a maximal queue length of $k$, and estimate the precision with Eq.~\eqref{eq:precision}.
Thus, we can merge $G_{\theta}(\mathbf{x}_t)$ and $P_{\theta}(\mathbf{x}_t, \mathbf{c})$ with respect to the control of precision with Eq.~\eqref{eq:Update-precision}. \gxz{Considering numerical stability, we normalize the inverse precision field \lms{in $\text{U}(\mathbf{x}_t)$} and clip the value in a fixed range. The formulation of $f$ is following:}
\begin{equation}
    f(\mathbf{w}) = \lambda \cdot \mathrm{clip}\left(\frac{\mathbf{w}-\Bar{\mathbf{w}}}{\sigma(\mathbf{w})}, 0, 1\right) + 1,
    \label{eq:Post-operation}
\end{equation}
where $\Bar{\mathbf{w}}$ and $\sigma(\mathbf{w})$ are the mean and standard error of $\mathbf{w}$, $\lambda$ is a constant that controls the guidance strength. Finally, we merge $G_{\theta}(\mathbf{x}_t)$ and $P_{\theta}(\mathbf{x}_t, \mathbf{c})$ with the \gxz{guidance} weight by inverse precision as Eq.~\eqref{eq:Update-precision}. The pseudo code of the inference process of CogDPM framework is shown in Algorithm~\ref{algo:inference-NPP}.

\paragraph*{\cky{Objective function.}}
CogDPM follows the training schema in diffusion probabilistic model~\cite{ho2020denoising} that predicts the noise from the corrupted inputs. We denote the loss term as $\mathcal{L}(\mathbf{\theta})$. The denoising U-Net  $\epsilon_\theta$ has parameters $\theta$, and takes the corrupted future observations $\mathbf{x}_{s}$, contexts $\mathbf{c}$ and the scalar diffusion step $s$ as input. We adopt the L1 loss to minimize the error between injected noise and the prediction of the denoising U-Nets. 

\begin{equation}
    \mathcal{L}(\mathbf{\theta})= \mathbb{E}_{t, \mathbf{x}_{0}, \epsilon, \mathbf{c}}\left[\Vert \mathbf{\epsilon} - \mathbf{\epsilon}_{\theta}\left(\sqrt{\bar{\alpha}_t}\mathbf{x}_{0} + \sqrt{1-\bar{\alpha}_t}\mathbf{\epsilon}, \mathbf{c}, t \right)\Vert_1\right]
\end{equation}

To jointly train the conditional and unconditional models,  $\mathbf{c}$ is replaced by $Z\sim \mathcal{N}(\mathbf{0},\mathbf{I})$ with 10\% probability. 

\begin{algorithm}[ht]
\caption{Inference Process of CogDPM framework}
\label{algo:inference-NPP}
\begin{algorithmic}
    \STATE {\bfseries Input:} Context input $\mathbf{c}$, denosing model $\epsilon_{\theta}$, maximul queue length $L$
    \STATE $\mathbf{x}_T \sim \mathcal{N}(\mathbf{0}_{\mathbf{x}}, \mathbf{I}_{\mathbf{x}})$
    \STATE Define free estimation queue $Q^{\text{free}}$
    \FOR{$t$ = $T$ \textbf{to} 1}
    \STATE $\mathbf{\epsilon}_{\mathbf{c}} \sim \mathcal{N}(\mathbf{0}_{\mathbf{c}}, \mathbf{I}_{\mathbf{c}})$ 
    \STATE $\epsilon_t^{\text{cond}} = \epsilon_\theta(\Hat{\mathbf{x}}_t,\mathbf{c})$ \algorithmiccomment{Network output with condition $\mathbf{c}$.}
    \STATE $\epsilon_t^{\text{free}} = \epsilon_\theta(\Hat{\mathbf{x}}_t,\epsilon_{\mathbf{c}})$ \algorithmiccomment{Network output without condition.}
    \STATE $P_\theta(\mathbf{x}_t,\mathbf{c})=
    \frac1{\sqrt{\alpha_t}}(\mathbf{x_t}-\frac{1-\alpha_t}{\sqrt{1-\bar{\alpha}_t}}\epsilon_t^{\text{cond}})$
    \STATE $G_\theta(\mathbf{x}_t)=\frac1{\sqrt{\alpha_t}}(\mathbf{x_t}-\frac{1-\alpha_t}{\sqrt{1-\bar{\alpha}_t}}\epsilon_t^{\text{free}})$
    \STATE $\Hat{\mathbf{x}}_{t\rightarrow 0} = \frac{1}{\sqrt{\bar{\alpha}_t}}\left(\mathbf{x}_t - \sqrt{1-\bar{\alpha}_t}\epsilon_t^{\text{free}}\right)$ \algorithmiccomment{Estimate $\mathbf{x}_0$ with $\mathbf{x}_t$.}
    \STATE Push $\Hat{\mathbf{x}}_{t\rightarrow 0}$ into $Q^{\text{free}}$
    \IF{Length of $Q^{\text{free}}$ exceeds $L$}
        \STATE Drop last term from $Q^{\text{free}}$ 
    \ENDIF
    \STATE Get inverse precision estimation $\mathbf{w} = f(\text{Var}(Q^{\text{free}}))$
    \STATE $\mathbf{x}_{t-1}=G_\theta(\mathbf{x}_t)+\mathbf{w}\cdot(P_\theta(\mathbf{x}_t,\mathbf{c})-G_\theta(\mathbf{x}_t))$ \algorithmiccomment{Prediction error minimization with precision weighting.}
\ENDFOR
\STATE {\bfseries Output:} $\mathbf{x}_0$
\end{algorithmic}
\end{algorithm}





\section{Experiments}

\begin{table*}[ht]
\centering
\label{tab:comparison_methods}
\begin{tabular}{@{}lcccccccc@{}}
\toprule
\multirow{2}{*}{Methods / Metrics} & \multicolumn{4}{c}{MovingMNIST}                           & \multicolumn{4}{c}{Turbulence}                             \\ 
                                   & \multicolumn{2}{c}{CRPS $\downarrow$}          & \multirow{2}{*}{\begin{tabular}[c]{@{}c@{}}CSI $\uparrow$ \\ \small{(w5)}\end{tabular}} & \multirow{2}{*}{\begin{tabular}[c]{@{}c@{}}FSS $\uparrow$\\ \small{(w5)}\end{tabular}} & \multicolumn{2}{c}{CRPS $\downarrow$}          & \multirow{2}{*}{\begin{tabular}[c]{@{}c@{}}CSI $\uparrow$\\ \small{(w5)}\end{tabular}} & \multirow{2}{*}{\begin{tabular}[c]{@{}c@{}}FSS $\uparrow$ \\ \small{(w5)}\end{tabular}} \\ \cmidrule(r){2-3} \cmidrule(r){6-7}
                                   & \small{(w8, avg)} & \small{(w8, max)} &                                                      &                                                     & \small{(w8, avg)} & \small{(w8, max)} &                                                      &                                                     \\ \midrule
FourCastNet                        & 0.0619            & 0.2288            & 0.1915                                               & 0.3261                                              & 0.0098            & 0.0119            & 0.3761                                               & 0.6558                                              \\
MotionRNN                          & 0.0377            & 0.1232            & 0.4859                                               & 0.6758                                              & 0.0037            & 0.0046            & 0.7235                                               & 0.9354                                              \\
PhyDNet                            & 0.0325            & 0.0983            & 0.6161                                               & 0.7969                                              & 0.0079            & 0.009             & 0.5456                                               & 0.8254                                              \\
PredRNN-v2                         & \textbf{0.027}             & 0.0774            & 0.688                                                & 0.8471                                              & 0.0033            & 0.0042            & 0.7529                                               & 0.9507                                              \\
DPM                            & 0.0323            & 0.082             & 0.6959                                               & 0.822                                               & 0.0023            & 0.0096            & 0.6725                                               & 0.9668                                              \\
CogDPM (ours)                    & \textbf{0.027}             & \textbf{0.0697}            & \textbf{0.7365}                                               & \textbf{0.8588}                                              & \textbf{0.0023}            & \textbf{0.0034}            & \textbf{0.7962}                                               & \textbf{0.9722}                                              \\ \bottomrule
\end{tabular}
\caption{Numerical Evaluation of Prediction Skills on MovingMNIST and Turbulence Datasets}
\end{table*}

\begin{figure*}[ht]
    \centering
    \includegraphics[width=\textwidth]{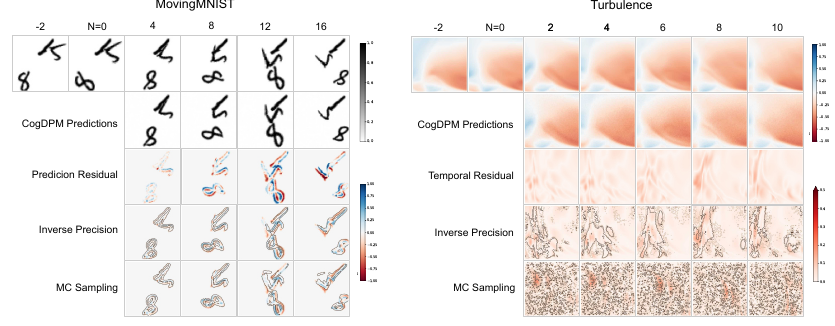}
    \vspace{-5mm}
    \caption{Predictions and inverse precision of CogDPM on rigid-body MovingMNIST dataset \textit{(left)} and Turbulence flow dataset \textit{(right)}.}
    \label{fig:CognitiveFeatures}
    \vspace{-5pt}
\end{figure*}

\begin{figure*}[ht]
    \centering
    \includegraphics[width=\textwidth]{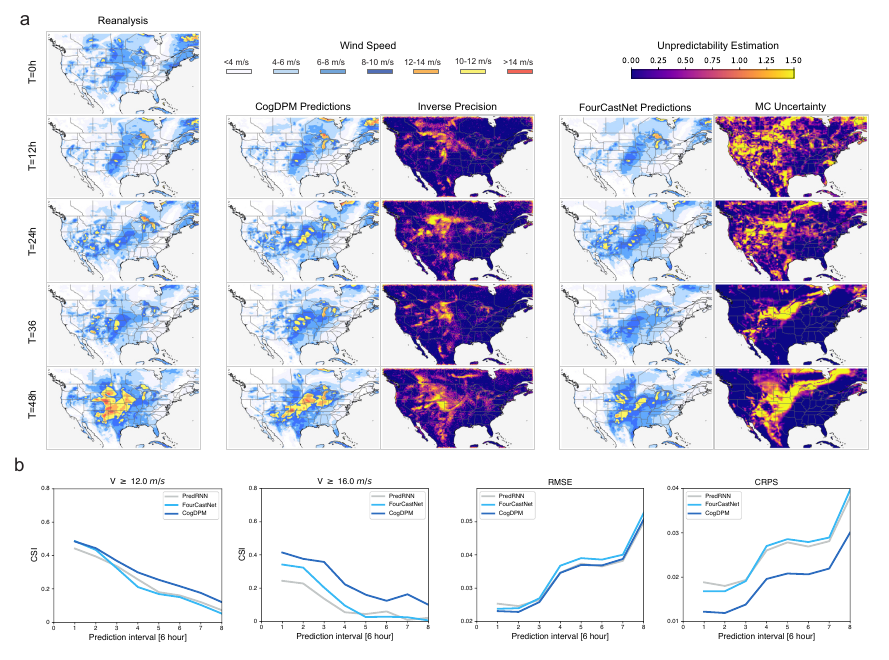}
    \vspace{0mm}
    \caption{\textbf{Experiments on high wind forecasting.} \textbf{a}, a Case study of the ERA5 wind forecast from 2017-03-04 18:00. High wind and tornadoes attacked the Mideast USA at 2017-03-06 18:00(T=48h)~\cite{Tornado06March2017}. CogDPM provides alarming forecasts, covering states with the most severe weather reports, Iowa and Missouri. CogDPM precision indicate the credibility of the predictions, helping forecasters to identify the missing and false positive regions. \textbf{b}, Numerical scores on ERA5 wind dataset from 2017-01-01 to 2019-12-31. We report CSI with 12 m/s (first) and 16 m/s (second) threshold, RMSE (third), and CRPS across four ensembles (fourth).}
    \label{fig:ERA5-case-metrics}
    \vspace{0pt}
\end{figure*}

\begin{figure}[ht]
    \centering
    \includegraphics[width=0.4\textwidth]{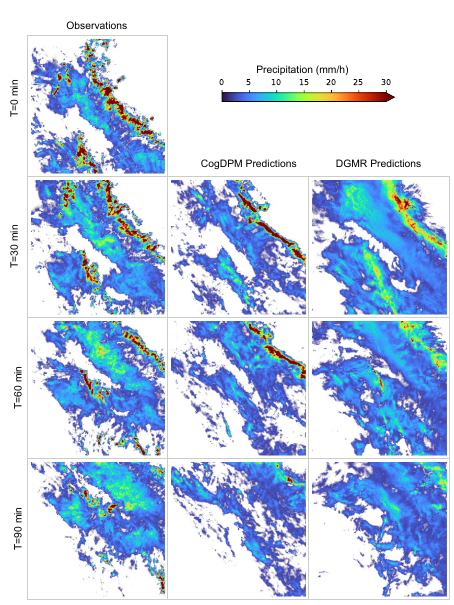}
    \vspace{-5pt}
    \caption{\textbf{Experiments on precipitation nowcasting. } Case study on an extreme precipitation event starting on 2019-07-24 at 03:15 in the UK timezone, CogDPM successfully predicts movement and intensity variation of the squall front, while DGMR produces results with early dissipation.}
    \label{fig:England-case-metrics}
    \vspace{-5pt}
\end{figure}

\begin{figure}[ht]
    \centering
    \includegraphics[width=0.5\textwidth]{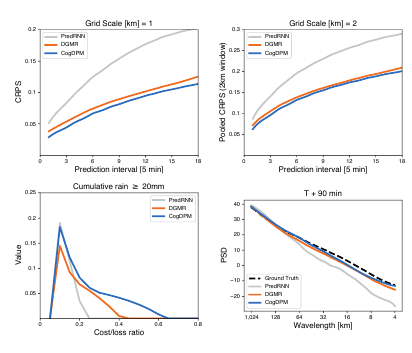}
    \vspace{-5pt}
    \caption{\textbf{Experiments on precipitation nowcasting. } Numerical verification scores on sampled the United Kingdom precipitation dataset in 2019. CRPS is computed with four ensembles for spatial pooling size 1km x 1km (left top) and 2 km x 2 km (right top); Economic value with 20 mm/h accumulative rain threshold (left bottom); Radially averaged power spectral density on predictions at 90 minutes (right bottom). CogDPM surpasses the operational forecast model DGMR in ensemble forecasting precision and forecast skillfulness.}
    \label{fig:England-case-metrics-2}
    \vspace{-5pt}
\end{figure}

We demonstrate that by incorporating the novel design inspired by the cognitive predictive process, CogDPM can deliver more skillful and improved results in tasks of scientific spatiotemporal field prediction.

\cky{\subsection{Synthesis Data Experiments}}
In this section, we compare the predictive performance of CogDPM with other mainstream deep predictive networks and investigate the interpretability of Precision weighting within the CogDPM framework in the context of spatiotemporal prediction. We expect high correlation between the precision estimation and the predictability of CogDPM. 
The inverse precision estimator should allocate more attention to the region with higher prediction difficulty.

\paragraph{Benchmarks.} We conduct experiments on the MovingMNIST dataset~\cite{wu2021motionrnn}, which simulates the motion of rigid bodies, and the Turbulence flow dataset, which models fluid dynamics.
The Moving MNIST dataset is generated with the same method as~\cite{wu2021motionrnn}. We create sequences with 20 frames, and each frame contains three handwriting digits. 
The motion of digits consists of transition, reflection, and rotation. Models predict the next 16 frames with 4 continuous context frames.
The turbulent flow dataset is proposed by~\cite{Wang2020TF}. We follow the same dataset parameters as~\citeauthor{Wang2020TF} and generate a sequence with 15 frames and 64 x 64 grids on each frame. Four frames are taken to predict the next 11 frames. 

\cky{We have selected a diverse array of deep spatiotemporal forecasting models as baselines for our study. These include the Transformer-based spatiotemporal forecasting model FourCastNet~\cite{pathak2022fourcastnet}
, RNN-type networks such as MotionRNN~\cite{wu2021motionrnn} and PredRNN-v2~\cite{wang2022predrnn}, the physics-inspired predictive model PhyDNet~\cite{guen2020disentangling}, and a predictive DPM model that employs naive Classifier-free Guidance~\cite{ho2021classifier-free} and utilizes the same network architecture as CogDPM.}

\cky{For the evaluation metrics, we have chosen the Neighborhood-based CRPS (Continuous Ranked Probability Score), CSI (Critical Success Index), and FSS (Fractional Skill Score), which are commonly used in scientific forecasting tasks. The CRPS metric emphasizes the ensemble forecasting capabilities of the model, with lower values indicating better predictive performance. On the other hand, the CSI and FSS metrics focus on assessing the accuracy of the model's predictions in peak regions, with higher values denoting stronger predictive capabilities. The implementation details of these metrics are provided in the appendix~\ref{appendix:metrics}, and we will continue to employ them in subsequent experiments on real-world datasets.}

\paragraph{Numerical Results} Table~\ref{tab:comparison_methods} presents the numerical evaluation results for two datasets. Here, $w$ denotes the window size employed in the Neighborhood-based assessment method, while $avg$ and $max$ represent the average and maximum values obtained from this method, respectively. The CogDPM model demonstrates consistent improvements over the baseline models in terms of the CRPS, which measures the average ensemble forecasting capability, as well as the CSI and FSS indicators, which assess the accuracy of the model's predictions in the peak regions. Additionally, when compared to the DPM model based on naive Classifier-free Guidance, CogDPM exhibits superior performance. This underscores the beneficial impact of introducing the Precision Weighting mechanism on enhancing the model's predictive efficacy.

\paragraph{Interpretability of precision weights.}
Figure~\ref{fig:CognitiveFeatures} presents the outcomes of the CogDPM model. The initial two rows delineate the ground truth images alongside the corresponding prediction results generated by CogDPM. 
The third row illustrates the prediction residuals, representing the discrepancies between the actual and predicted data as depicted in the preceding rows. The fourth row features images that overlay the inverse precision map, highlighting the top 20\% of values with a black contour line, against a backdrop of the residual map. \cky{The fifth row shows the precision map estimated by Monte Carlo sampling which estimate the prediction confidence with the variation among multiple independent predictions with difference noise prior~\cite{zhang2021modern}.}

CogDPM provides reasonable predictions in both datasets. 
In the prediction of rigid body motion, the estimated Inverse Precision effectively encompasses the Precision Residuals, which are primarily located at the edges of objects. 
The edges of objects present a greater challenge for prediction compared to blank areas or the interior of objects. This outcome aligns with our expectations for the estimation of the precision map.
\cky{Precision estimated with MC sampling works similarly but provide more false positive region in frame 12 and 14.}

In the prediction of fluid motion, regions with large temporal residuals exhibit higher accelerations, indicating increased predictive difficulty. 
The estimated Inverse Precision indeed covers the Temporal Residuals well, meeting our expectations. 
We observe that in both fluid and rigid body motion prediction tasks, the Precision weights of CogDPM exhibit varying styles, yet consistently depict the model's confidence on current case.
\cky{On comparison, MC sampling method almost fails in this case due to the over-confidence of the prediction result. Difference among multiple predictions have no significant signals but random noise. While, the CogDPM is not effected because its precision describe the continuous enhancing process of model's confidence during the hierarchy inference.}

\subsection{Surface Wind Forecasting Experiments}

\textbf{Benchmarks.} We first evaluate our model by applying it to the task of surface wind forecasting, using the ERA5 reanalysis dataset~\cite{era5}. 
Accurate wind field forecasting is crucial for various applications in energy and weather domains. 
Ensemble forecasting is a key technique to provide more useful information for the forecasters, which provides multiple predictions and the confidence of its predictions.
We show that CogDPM not only provide better ensemble forecasts results, but also estimate the prediction confidence with its precision weights.

We choose real-world operational metrics for evaluation. 
In the meteorology domain, forecasters focus on evaluating the risk of high wind and confirming the time for extreme weather issue warnings.
On this purpose, we use Critical Success Index (CSI) to measure the consistency between heavy wind regions in forecasts and ground truths. 
In the energy domain, accurate wind field forecasting supports the prediction of wind power, which is essential for the fluctuation control of clean energy generation~\cite{marugan2018survey}. 
Absolute wind speed is the dominant factor that affect the power production of the wind turbine~\cite{porte2013numerical}; thus, we consider pixel-wise Root Mean Square Error (RMSE) and Radially Continuous ranked probability score (CRPS) on wind speed for the evaluation of this scenario~\cite{barbounis2006long}.
Applendix~\ref{appendix:metrics} shows detailed implemation of these metrics.

\textbf{Results.} We use the ERA5 reanalysis surface wind data and crop patches centered in the US spanning from 1979 to 2021. 
We evaluate predictions for the next 48 hours with 6-hour intervals using the observations in past 24 hours. 
We compare the proposed method with FourCastNet~\cite{pathak2022fourcastnet}, a domain-specialized network for reanalysis field forecasting, and predictive recurrent networks for deterministic video prediction. 
FourCastNet provides ensemble forecasts based on the Gaussian disturbance on the initial states following~\cite{Ensemble2003geir}. 
%

Figure~\ref{fig:ERA5-case-metrics}a shows studies on a case starting from 2017-03-04 18:00. 
The results from FourCastNet indicate a failure to accurately forecast the growing high wind region, and the high wind region is underestimated in the 48-hour forecast. 
In contrast, results from CogDPM not only locate the high wind region more accurately, but also provide intensity estimates much closer to the ground truth, supporting the need for 48-hour-ahead precautions.
CogDPM are capable of providing alarming forecasts around 2017-03-06 18:00, when high wind and tornadoes attacked the Mideast USA\footnote{Summary of March 06 2017 Severe Weather Outbreak - Earliest Known Tornado in Minnesota's History, \url{https://www.weather.gov/mpx/SevereWeather\_06March2017}}.

We also visualize the inverse precision fields corresponding to the forecasts, since confidence estimation provide key information for decision-making. 
In the forecast for the first 24 hours, the uncertainty fields given by FourCastNet are relatively dispersed and not closely related to the evolution of the associated wind field. 
In the next time period to the 48 hours, FourCastNet produces unreasonable estimates for the windless area in the upper right corner.
The inverse precision fields given by CogDPM had much closer correlations to the weather process. 
In the 48-hour forecast, CogDPM underestimated the forecast intensity in Wyoming and Colorado, but allocated lower precision on that region.

Figure~\ref{fig:ERA5-case-metrics}b shows that CogDPM outperforms baseline methods on CSI, particularly for heavier wind thresholds. 
For the measurement of RMSE, we take the mean across eight ensemble forecasts for all methods. 
Although DPMs are not directly optimized by the Mean Squared Error (MSE) loss, the mean ensemble results are competitive with predictive models trained with MSE losses. 
The CogDPM exhibits a lower CRPS across all prediction times, indicating its ability to effectively generate ensemble forecasts.

Our results demonstrate that CogDPM is capable of making predictions under severe conditions, supported by the probabilistic forecast ability of the PEM process, while deterministic models avoid predicting severe cases to reduce mistake-making risk.

\subsection{Precipitation Nowcasting Experiments}

\textbf{Benchmarks.} We evaluate our model on the precipitation nowcasting task using the United Kingdom precipitation dataset~\cite{ravuri2021skilful}.
Precipitation nowcasting aims to predict high-resolution precipitation fields up to two hours ahead, which provides socioeconomic value on weather-dependent decision-making~\cite{ravuri2021skilful}. 
Precipitation data is extremely unbalanced on spatiotemporal scales, demanding nowcasting models to focus on vital parts of the field. 
Fig.~\ref{fig:England-case-metrics}a shows a case study selected by the chief meteorologist from MetOffice~\cite{ravuri2021skilful}, which involves a squall line sweeping across the United Kingdom. 
We choose DGMR as a strong baseline on skillful nowcasting~\cite{ravuri2021skilful},
which is data-driven method that forecast precipitation with a generative adversarial network. 
DGMR is also the operational method deployed by Met Office of the United Kingdom.

\textbf{Results.} In Figure~\ref{fig:England-case-metrics}, our results accurately forecast both the trajectory and intensity fluctuations of the squall line, as depicted by the red precipitation line in the top right segment. 
CogDPM's forecasts consistently show the squall line progressing over 30 and 60 minutes, followed by dissipation at the 90-minute mark, mirroring actual events.
Conversely, predictions from DMGR indicate a rapid dissipation of the squall line within 30 minutes, and significantly weaker outcomes are projected for the 60-minute mark.
We posit that the suboptimal performance of the DGMR model is attributable to the simultaneous use of generative loss and pixel-wise alignment loss functions during its training phase, which leads to unstable training process and still keeps the drawback of dissipation of deterministic alignments. 
While the generative loss alone is capable of simulating realistic meteorological processes, it falls short in accurately predicting the extent of precipitation and is abandoned in DGMR. 
%
%
On the contrary, CogDPM does not require additional deterministic alignment during training but enhances precision with precision-weighted guidance during inference steps.
We present additional case studies in Appendix~\ref{appendix:additional_case_studies_2}.

We further explore the numerical evaluations in Fig~\ref{fig:England-case-metrics-2} with metrics on different forecast properties focusing on the accuracy, reality and diversity. 
Radially Continuous ranked probability score (CRPS) measures the alignment between probabilistic forecast and the ground truth. 
We also report the spatially aggregated CRPS~\cite{ravuri2021skilful} to test prediction performance across different spatial scales. Details of these metrics can be found in Extended Data. The first row in Fig ~\ref{fig:England-case-metrics} shows CogDPM consistently outperforms baseline models for the whole time period. 
We adopt the decision-analytic model to evaluate the Economic value of ensemble predictions~\cite{ravuri2021skilful}. Curves in \cky{Figure~\ref{fig:England-case-metrics-2}} with greater under-curve area provide better economic value, and CogDPM outperforms baseline models in this regard. 
Radially averaged power spectral density (PSD) evaluates the variations of spectral characteristics on different spatial scale. CogDPM achieves the minimal gap with ground truth characteristics. 

The superior performance metrics of CogDPM stem from its diffusion models' ability to emulate the hierarchical inference of predictive coding, resulting in smaller prediction errors compared to single-step forecasting models. 
Furthermore, the integration of precision weighting allows the model to dynamically assess the precision of inputs and adjust the intensity of conditional control accordingly. This targeted approach effectively reduces errors in areas that are challenging to predict, thereby enhancing the accuracy of the model in delineating boundaries and extreme regions.

\section{Discussion}
CogDPM is related to classifier-free diffusion models~\cite{ho2021classifier-free}, which enhance the class guidance with a conditional DPM and an unconditional DPM. 
CogDPM framework builds the connection between classifier-free diffusion models and predictive coding.
We also introduce the precision estimation method with the reverse diffusion process and use precision to control the guidance strength in spatiotemporal scales.
%
We adopt the ablation study to show the enhancement in prediction skills of the CogDPM framework compared with the vanilla CFG method in appendix~\ref{appendix:ablantion_study}.

\textit{Active inference}~\cite{parr2019perceptual} is also a widely discussed theory of the predictive coding framework, which states that cognition system actively interact with the environment to minimize the prediction error.
Active inference is omitted in this work. We take a computational predictive coding model with both active inference and precision weighting as the future work.

\section{Conclusion}
We propose CogDPM, a novel spatiotemporal forecasting framework based on diffusion probabilistic models. 
%
CogDPM shares main properties with predictive coding and is adapted for field prediction tasks.
The multi-step reverse diffusion process models the hierarchy of predictive error minimization. 
The precision of a latent expectation can be estimated from the variance of states in the neighboring levels.
The CogDPM framework has demonstrated its ability to provide skillful spatiotemporal predictions in precipitation nowcasting and wind forecasting. Case studies and numeric evaluations demonstrate that CogDPM provides competitive forecasting skills.
%

\section*{Impact Statements}
This paper presents work whose goal is to advance the deep learning research for a PC-based spatiotemporal forecasting framework. There are many potential societal consequences of our work, none of which we feel must be specifically highlighted here.

\bibliography{sn-bibliography.bib}
\bibliographystyle{icml2024}

\newpage
\appendix
\onecolumn
\section{Preliminary}

\paragraph{Diffusion Models.}
Diffusion models are the state-of-the-art deep generative models on image synthesis\cite{dhariwal2021diffusion,song2019generative,ho2020denoising}, and have been explored widely in numerous tasks, such as computer vision\cite{baranchuk2021label}, time series modeling~\cite{rasul2021autoregressive} and molecular graph modeling~\cite{hoogeboom2022equivariant,xu2021geodiff}. 
Diffusion probabilistic models (DPMs), a major paradigm in diffusion models, handly construct the forward process $q(\mathbf{x}_{1:T}\mid \mathbf{x}_0)$ by progressively injecting noise to a data distribution $q(\mathbf{x}_0)$, and generate samples with a denoising backward process. Formally, we define a Markov forward process $q$ with latent variables $\mathbf{x}_1, \mathbf{x}_2, \dots \mathbf{x}_T$, which follows Eq.~\eqref{Eq:Forward1} and Eq.~\eqref{Eq:Forward2}:

\begin{equation}
    q(\mathbf{x}_{1:T}\mid \mathbf{x}_0)= \prod_{t=1}^T q(\mathbf{x}_{t}\mid \mathbf{x}_{t-1}),
    \label{Eq:Forward1}
\end{equation}

\begin{equation}
    q(\mathbf{x}_t\mid \mathbf{x}_{t-1}) = \mathcal{N}(\mathbf{x}_t\mid \sqrt{1-\beta_t} \mathbf{x}_{t-1}, \beta_t \mathbf{I}),
    \label{Eq:Forward2}
\end{equation}

where $\beta_t \in (0,1),\ t=1,\dots, T$ schedule the forward process. In this work, we select cosine $\beta$ scheduling~\cite{nichol2021improved}.
Eq.~\eqref{Eq:Forward3} allows to directly sample an arbitrary latent variable conditioned on the input $\mathbf{x}_0$. Let $\alpha_t = 1-\beta_t$ and $\bar{\alpha}_t = \prod_{s=1}^{t} \alpha_s$, we formulate the marginal distribution as:

\begin{equation}
    q(\mathbf{x}_t\mid \mathbf{x}_0) = \mathcal{N}(\mathbf{x}_t\mid \sqrt{\bar{\alpha}_t} \mathbf{x_0} , (1-\bar{\alpha}_t) \mathbf{I}).
    \label{Eq:Forward3}
\end{equation}

We define the reverse process for Eq.~\eqref{Eq:Forward1} and Eq.~\eqref{Eq:Forward2} as $p_\theta (\mathbf{x}_{0:T})$, with initial state $p(\mathbf{x}_T) = \mathcal{N}(\mathbf{x}_T\mid \mathbf{0}, \mathbf{I})$ and parameterized marginal distributions:
\begin{equation}
    p_{\theta}(\mathbf{x}_{0:T}) = p(\mathbf{x}_T)\prod_{t=1}^T p_{\theta}(\mathbf{x}_{t-1}\mid \mathbf{x}_{t}),
    \label{Eq:reverse-fractorization}
\end{equation}
\begin{equation}
    p_{\theta}(\mathbf{x}_{t-1}\mid\mathbf{x}_{t}) = \mathcal{N}(\mathbf{x}_{t-1} \mid \mu_{\theta}(\mathbf{x}_t, t), \Sigma_{\theta}(\mathbf{x}_t, t)).
    \label{Eq:reverse-marginal-dist}
\end{equation}

Diffusion models transfer the goal of generating target distribution into minimizing distance between forward and backward processes, formulated in Eq.~\eqref{eq:kl_forward_backward}. Eq.~\eqref{eq:marginal_align} shows that the alignment between two processes can be factorized into the alignment between marginal conditional distributions. 

\begin{align}
    &\min_{\{\mathbf{\mu}_t, \Sigma_t^2\}_{t=1}^T} 
    L_{\text{vb}}(q,p_{\theta}) \\
    \Leftrightarrow& \min_{\{\mathbf{\mu}_{t}, \Sigma_{t}^2\}_{t=1}^T} D_{\text{KL}}(q(\mathbf{x}_{0:T})\Vert p_{\theta}(\mathbf{x}_{0:T})) \label{eq:kl_forward_backward}\\
    \Leftrightarrow& \min_{\{\mathbf{\mu}_t, \Sigma_t^2\}_{t=1}^T}
    \sum_{t=1}^T D_{\text{KL}}(q(\mathbf{x}_{t-1}\mid \mathbf{x}_{t}, \mathbf{x}_0)\Vert p_{\theta}(\mathbf{x}_{t-1}\mid \mathbf{x}_{t})).
    \label{eq:marginal_align}
\end{align}
Ho et al. \cite{ho2020denoising} adopt a denoising network $\mathbf{\epsilon}_{\theta}(\mathbf{x}_t, t)$ to parameterize $\mu_{\theta}(\mathbf{x}_t, t)$, and simplify the above loss as Eq.~\eqref{Eq:simplied_loss}:
\begin{equation}
    \mathrm{E}_{t\sim \mathcal{U}(1,T), \mathbf{x_0}\sim q(\mathbf{x}_0)\mathbf{\epsilon}\sim \mathcal{N}(\mathbf{0}, \mathbf{I})}\Vert \mathbf{\epsilon} - \mathbf{\epsilon}_{\theta}(\mathbf{x}_{t}, t)\Vert^2.
    \label{Eq:simplied_loss}
\end{equation}
We train the diffusion models with Eq.~\eqref{Eq:simplied_loss}, and CogDPM inference results with the same methodology as Eq.~\eqref{Eq:reverse-marginal-dist}.

\paragraph{Video generation with Diffusion Models.}
Considering the success in image synthesis, diffusion models can be naturally applied to video generation and prediction tasks. Unlike static images, video generation faces two main problems: 1. considerable computation consumption, which also cause slow inference speed; 2. inconsistency between adjacent frames. 
For the first problem, previous works focus on enhancing the computation efficiency of the back-bone U-Net in diffusion models \cite{ho2022video,voleti2022mcvd}. 
Video-Image joint training helps accelerating the optimization progress \cite{ho2022video}.

\citet{yang2022diffusion} combine a deterministic prediction model with a residual prediction diffusion model. The diffusion model learn to generate the stochastic error between the deterministic prediction and the ground truth video, and the deterministic model ensure the continuity between adjacent frames. \citet{ho2022video} introduce a gradient based conditional sampling method to improve temporally coherency, and also adapt cascaded architectures from image-based to video-based diffusion models\cite{ho2022imagen}, leading to enhanced video quality.

To maintain temporal consistency across extended video sequences, researchers also design auto-regressive conditioning procedures \cite{voleti2022mcvd,harvey2022flexible,hoppe2022diffusion,yang2022diffusion}. These models recursively utilize preceding outputs as inputs to sequentially generate subsequent frames. Voleti et al. and Höppe et al. use masked sequences to train the model\cite{voleti2022mcvd,hoppe2022diffusion}, while
Yang et al. treat the process from a purely probabilistic view\cite{yang2022diffusion}.

\citet{Blattmann_2023_CVPR} extend 2D Latent Diffusion Models (LDMs) to 3D versions by inserting temporal layers between each blocks, which called spatial layers, in original U-Nets. Spatial layers concentrate on synthesizing individual frames in the video, while temporal layers are dedicated to the alignment between different frames. Leveraging pre-trained 2D LDMs, 3D LDMS can generate long videos without losing much image quality. Stable Video Diffusion is a large-scale implementation of 3D LDMs \cite{blattmann2023stable}, which exhibits its outstanding performance in video sythesis.

Previous works prove that DPMs can generate realistic and coherent videos with considerable diversity, and corresponding experiments mainly focus on general videos photoed by RGB cameras.
Spatiotemporal forecasting is another brand of video prediction, which focus on scientific applications. In these tasks, beyond frame consistency, forecast accuracy, diversity and skillfulness are the mainly concerned metrics. In this work, we explore the field evolution forecasting with diffusion models, and enhance model forecasting value on real-world applications.

\section{Implementation Details}

\paragraph*{Architecture design.} 
CogDPM adopts the U-Net~\cite{unet2015icm,ho2022video} backbone coupled with a vision transformer (ViT)~\cite{dosovitskiy2020image} encoder. 
 
Inputs of the ViT encoder contain three parts: 1) patches of context cubes, 2) cube positional embedding, and 3) diffusion step embedding. The inputs are summed together after a linear projection and then fed into the ViT encoder. We adopt a random mask among the ViT inputs for better efficiency. 

The architecture of the U-Net consists of a down-sampling tower, mid-blocks, an up-sampling tower, and short-cut connections. The down-sampling tower progressively reduces the spatial dimensions of the input while increasing the number of channels. The mid-blocks maintain the shape of the intermediate representations. The up-sampling tower reverses the operations of the down-sampling tower, and the short-cut connections provide direct connections between corresponding down-sampling and up-sampling blocks.

A single U-Net block employs separate neural modules for spatial and temporal modeling, including a spatial residual block with convolution neural networks and a temporal block with axial attention. We use cross-attention layers to merge the representations from the context encoder into the U-Net blocks. We repeat these blocks several times, followed by a bilinear interpolation operation that doubles or halves the spatial shapes.

We further introduce the cascade diffusion pipeline to accelerate sampling speed for the participation nowcasting task~\cite{ho2021cascaded}. We parallel couple a low resolution CogDPM model with an additional super-resolution CogDPM, and allocate fewer inference steps on the super-resoluation model to decrease the total inference time compared to a single high-resolution model.

\paragraph*{Evaluation.}
For the precipitation nowcasting, we follow the metrics used in~\cite{ravuri2021skilful}, including CRPS, window pooled CRPS, economic values and PSD. The computational details are listed in the supplementary materials. 

We uniformly sample 10,000 cases from the 512 km $\times$ 512 km cropped test dataset provided by~\cite{ravuri2021skilful}. We crop the central 256 km $\times$ 256 km as model inputs to keep the same input shape with the pretrained DGMR. We take quantitative evaluations of the central 128 km $\times$ 128 km to avoid the boundary effects, as the similar method in~\cite{ravuri2021skilful}.  

For surface wind forecasting, we report CSI for the high wind precaution task and report CRPS and RMSE for the electric power prediction task. 

We also evaluate on the central 64 $\times$ 144 grids to concentrate on the land measurement while also eliminating the boundary effect from partial observation.

\paragraph*{Training.}
We train a two-stage cascade diffusion model for the United Kingdom precipitation dataset and one-stage diffusion models for the others.

The one-stage models are trained for $1 \times 10^6$ steps. The learning rate is $2 \times 10^{-5}$, using Adam optimizer with $\beta_{1} = 0.9$ and $\beta_2 = 0.999$. We randomly replace conditions as i.i.d. standard Gaussian noise with a probability of 10\%, following the classifier-free diffusion models~\cite{ho2021classifier-free}. 
On the ERA5 dataset, we train the model on 2 GPU cores (NVIDIA A100) for two weeks using a batch size of 16 per training step. 
On the turbulence flow and Moving MNIST dataset, we train the model on 1 GPU core (NVIDIA A100) for one week using a batch size of 36 per training step. 

The two-stage model cascades a low-resolution DPM and a super-resolution DPM. 
The low-resolution DPM has the same training recipe as a model for Moving MNIST. 
The super-resolution DPM is trained for $5 \times 10^6$ steps, using 4 GPU cores (NVIDIA A100) for two weeks using a batch size of 8 per training step. 

\section{Datasets}
\label{appendix:datasets}
In this study, we conduct experiments on synthesis datasets for interpreting precision estimations and on real-world datasets for evaluating prediction skills.

For interpreting precision estimations, we adopt The Moving MNIST dataset which is generated with the same method as~\cite{wu2021motionrnn}. We create sequences with 20 frames, and each frame contains three handwriting digits. 
The motion of digits consists of transition, reflection, and rotation.
We use four initial frames to predict the movements of the digits in the following 16 frames, and each frame has 64 $\times$ 64 grids.
We generate 100,000 sequences for training, 1,000 for validation and 1,000 for testing. 
The turbulent flow dataset is proposed by~\cite{Wang2020TF}. We follow the same dataset parameters as~\cite{Wang2020TF} and generate a sequence with 15 frames and 64 x 64 grids on each frame. Four frames are taken to predict the next 11 frames. We generate 20,000 sequences for training, 1,000 for validation, and 1,000 for testing. 

For testing the wind field forecasting, we use the ERA5 dataset, a high-resolution global atmospheric reanalysis produced by the European Centre for Medium-Range Weather Forecasts (ECMWF)~\cite{era5}. 
We select the region covering the United States, the region from longitude 130 degrees West to 60 degrees West, and latitude 20 degrees North to 56 degrees North. The dataset has the time scale from 1959 to 2019, and the 6-hour time interval. 
We use the 24-hour surface wind and zonal wind speed to predict the next 48 hours. 
We use the data from 1959-01-01 to 2013-12-31 for training, 2014-01-01 to 2016-12-31 for validation, and 2017-01-01 to 2019-12-31 for testing. 

For evaluating the skill of precipitation nowcasting of CogDPM, we adopt the United Kingdom precipitation dataset, which contains radar composites from the Met Office RadarNet4 network from 2017 to 2019, and the experiment settings used in~\cite{ravuri2021skilful}.
The dataset is patched into 24 $\times$ 256 $\times$ 256 composites with a time interval of 5 minutes and 1 km x 1 km spatial grids. 
The model forecasts precipitation fields of 90 minutes with 20-minute observations.
\cky{We note that the UK precipitation dataset is a large high resolution spatiotemporal forecasting dataset which takes about 1 TB saved with TF Records.}
We follow the dataset splits and importance sampling techniques described in~\cite{ravuri2021skilful}. 

We list the detailed information of these datasets in the table~\ref{tab:datasets-info}.

\begin{table}[htbp]
\label{tab:datasets-info}
\centering
\caption{Overview of Datasets for Different Tasks}
\label{tab:datasets}
\begin{tabular}{@{}|c|c|c|c|c|c|@{}}
\hline
\textbf{Feature} & \textbf{MovingMNIST} & \textbf{Turbulence} & \textbf{US Surface wind} & \textbf{UK Precipitation} \\ 
\hline
Image shape & (64, 64) & (64, 64) & (144, 280) & (256, 256) \\ 
Sequence length & 20 & 15 & 12 & 22 \\ 
Channel & 1 & 2 & 3 & 1 \\ 
Size of training set & 100,000 & 100,000 & 73,000 & 5,788,800 \\ 
Size of validation set & 1,000 & 1,000 & 4,380 & 1,000 \\ 
Size of test set & 1,000 & 1,000 & 4,380 & 10,000 \\ 
\bottomrule
\end{tabular}
\end{table}

\section{Verification Metrics}
\label{appendix:metrics}
We outline the five standard evaluation metrics used in this article.

\textbf{Critical Success Index (CSI)}~\cite{cite:csi1990} quantifies the accuracy of binary predictive decisions, determining whether the target intensity value exceeds a specific threshold. To calculate this metric, we sum the hits, misses, and false alarms across all grid points and compute their ratio. CSI evaluates precision and recall simultaneously and is widely used to assess high wind forecasting. It is worth noting that the CSI metric counts the hit, miss, and false alarm over the entire test set.

\textbf{CSI-Neighborhood}~\cite{cite:csiwindow2012} is the CSI metric computed based on the neighborhood. Neighborhood methods evaluate the forecasts within a spatial window surrounding each grid, which can assess the `closeness' of the forecasts. This metric is particularly suitable for verifying high-resolution forecasts. 

\textbf{Radially Continuous ranked probability score (CRPS)} ~\cite{crps2007} measures the alignment between probabilistic forecasts and ground truth data. It is widely used in the evaluation of weather forecasting models. The CRPS takes into account the entire distribution of forecasting probabilities, making it a more comprehensive evaluation metric compared to traditional point forecast metrics.
We also report the neighborhood CRPS~\cite{ravuri2021skilful}, also following~\cite{cite:csiwindow2012}

\textbf{Fractional Skill Score (FSS)} as described by Roberts and Lean (2008)~\cite{roberts2008scale} represents another neighborhood-based metric, delineated by target thresholds. For each grid cell within a piece of test data, the proportion of surrounding cells exceeding a defined threshold within a spatial window is calculated. Subsequently, the summation of these proportions—predicted versus observed—across all grid cells is termed the Fractional Brier Score (FBS). The FSS is derived from the normalized FBS, relying on a threshold to determine local value occurrences and employs the Fractional Brier Score (FBS) to contrast predicted and observed value frequencies. Unlike the CSI-Neighborhood, which solely focuses on the accuracy of hit predictions, the FSS also facilitates the comparison of the rate of grid cells exceeding the threshold within a spatial window.

\textbf{Power Spectral Density (PSD)}~\cite{psd2001,psddetail2005} measures the power distribution over each spatial frequency, comparing the intensity variability of forecasts to that of the observations. We use the PSD implementation from the PySTEPS package~\cite{pysteps2019gmd}. Forecasts that have minor differences with observations are preferred.

\textbf{Economic Value} \cite{ravuri2021skilful} evaluates the outcome of forecasts with a cost-loss ratio decision model. It shows the relative loss decrease with a forecasting-based precaution policy for a particular cost-loss ratio. For fair comparison, we follow the implementation from DGMR~\cite{ravuri2021skilful}.

\section{Ablation Study}
\label{appendix:ablantion_study}
\begin{figure}
    \centering
    \includegraphics[width=\textwidth]{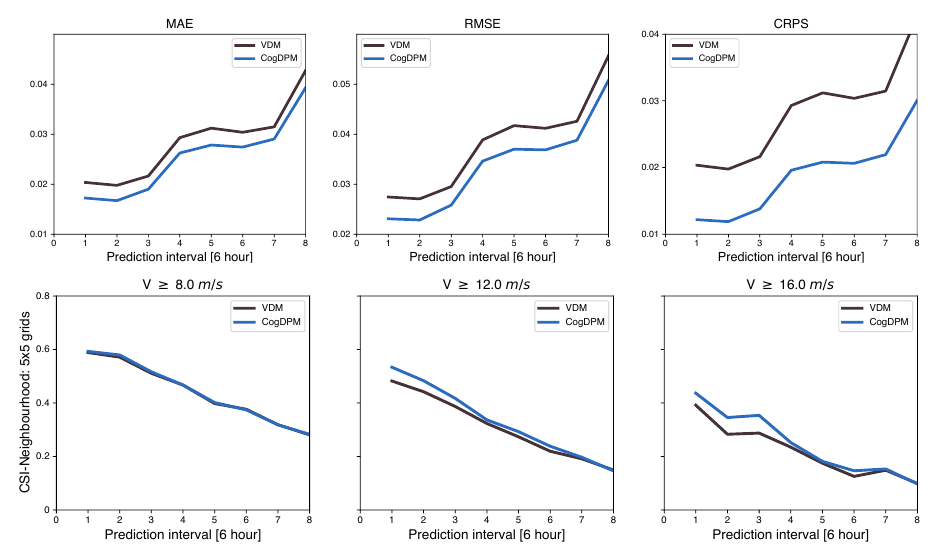}
    \caption{\textbf{Numerical comparison between CogDPM and video diffusion models (VDM) on the ERA5 wind forecast task.} The first row shows CSI metrics with thresholds of 8.0 m/s, 12.0 m/s and 16.0 m/s. The second row shows MAE, RMSE and CRPS relatively.}
    \label{fig:AblationERA5metrics}
\end{figure}

We demonstrate ablation experiments on precision weighting mechanism and network design. 
We compare the video diffusion model (VDM)~\cite{ho2022video} with numerical evaluation. VDM shares the same training schema and U-Net architecture with CogDPM, and maintains a constant Classifier-free guidance during inference.
Figure~\ref{fig:AblationERA5metrics} reports the performance of MAE, RMSE, CRPS, and CSI metrics with different wind speed thresholds on the ERA5 surface wind dataset. The results show that CogDPM has significant enhancements in MAE, RMSE, and CRPS metrics, and achieves comparable or better results on CSI indices with different wind speed thresholds. This demonstrates that precision-weighted guidance can increase the precision of CogDPM in high-wind regions and provide more diverse forecasts, thereby improving the skillfulness of the forecast results.

\section{Additional Case Studies}
\label{appendix:additional_case_studies_2}

\begin{figure}
    \centering
    \includegraphics[width=\textwidth]{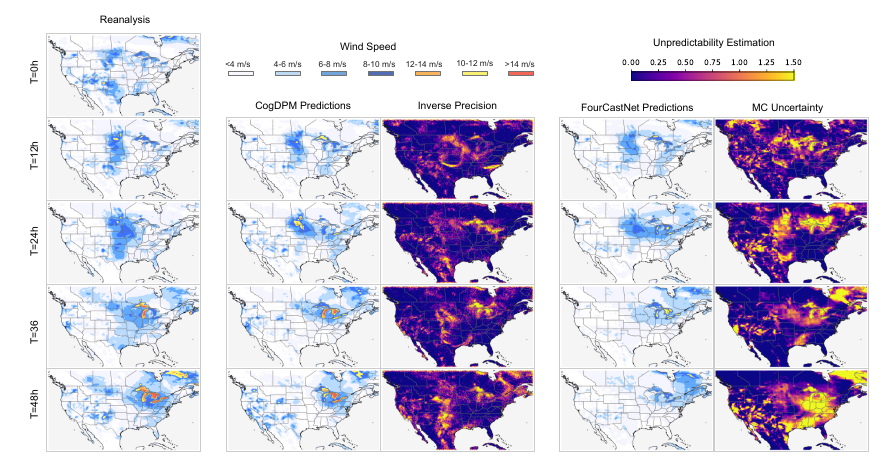}
    \caption{\textbf{Additional case studies on ERA5 surface wind prediction.} \textbf{a} Case study of the ERA5 wind forecast from 2017-01-02 18:00. The CogDPM successfully prediction the high wind region moving from mideast USA to the Great Lakes. FourCastNet overestimate the moving spatial scale and underestimate its intensity for T=36h and T=48h. The inverse precision field of CogDPM indicate an unpredictable region for mideast USA at T=48h where the prediction neglect. FourCastNet uncertainty focus on the east-south USA at T=48, but is irrelevant with the truth field. Maps produced by the Cartopy package.}
    \label{fig:ERA5extraCase}
\end{figure}

In Figure~\ref{fig:ERA5extraCase}, we present additional case studies on ERA5 surface wind prediction. The inverse precision field of CogDPM indicate an informative unpredictable region.

\begin{figure}
    \centering
    \includegraphics[width=0.5\textwidth]{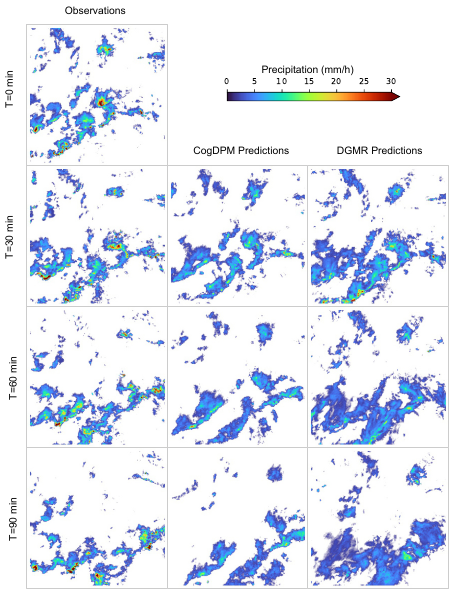}
    \caption{\textbf{Additional case studies on the United Kingdom precipitation dataset.} In this case, the forecast results of CogDPM maintained the bifurcated structure of the two squall lines and their precipitation range, with predicted locations closely matching actual observations. On the other hand, the results of DGMR showed the two squall lines merging at 60 minutes, and the 90-minute forecast significantly misreported the precipitation range. The precision field of CogDPM effectively described the boundaries of the precipitation range, while the precision field of DGMR duplicated the predicted precipitation intensity information.}
    \label{fig:EnglandExtraCase}
\end{figure}

In Figure~\ref{fig:EnglandExtraCase}, we present additional case studies on the United Kingdom precipitation dataset. The precision field of CogDPM effectively described the boundaries of the precipitation range. 

\end{document}